%% file: acl2023.tex
\pdfoutput=1

\documentclass[11pt]{article}

\usepackage{ACL2023}

\usepackage{times}
\usepackage{latexsym}

\usepackage[T1]{fontenc}

\usepackage[utf8]{inputenc}

\usepackage{microtype}

\usepackage{inconsolata}

\usepackage{times}
\usepackage{latexsym}
\usepackage{hyperref}
\usepackage{amsmath}
\usepackage{graphicx}
\usepackage{booktabs}
\usepackage{multirow}
\usepackage{amssymb}
\usepackage{pifont}
\usepackage{color,soul}
\usepackage{pifont}
\usepackage{microtype}
\usepackage{graphicx}
\usepackage{subfigure}
\usepackage{booktabs} 
\usepackage{hyperref}
\usepackage{url}
\usepackage{multirow}
\usepackage{wrapfig}
\usepackage{enumitem}
\usepackage{times}
\usepackage{latexsym}
\usepackage{amsmath}
\usepackage{amsthm}
\usepackage{amssymb}
\usepackage{xspace}
\usepackage{amsfonts,eucal,amsbsy}
\usepackage[normalem]{ulem}
\usepackage{sidecap}

\sidecaptionvpos{figure}{t}
\input{math_commands.tex}

\newcommand{\model}{\textsc{VMSST}\xspace}
\newcommand{\modelfull}{Variational Multilingual Source-Separation Transformer\xspace}

\newcommand{\N}{92\xspace}
\newcommand{\contrastive}{\textsc{Contrastive}\xspace}
\newcommand{\bitrans}{\textsc{BiTranslation}\xspace}
\newcommand{\mgtcontrastive}{\textsc{VMSST Contrastive}\xspace}

\newcommand{\titlestr}{Beyond Contrastive Learning: A Variational Generative Model for Multilingual Retrieval}
\title{\titlestr}

\author{John Wieting$^1$, Jonathan H. Clark$^1$, William W. Cohen$^1$, \\
{\bf Graham Neubig}$^2$, \bf{and Taylor Berg-Kirkpatrick}$^3$ \\
  $^1$Google DeepMind \\
  $^2$Carnegie Mellon University,
  Pittsburgh, PA, 15213, USA \\
  $^3$University of California San Diego,
  San Diego, CA, 92093, USA\\
  {\small \texttt{\{jwieting,jhclark,wcohen\}@google.com}, \texttt{gneubig@cs.cmu.edu}, \texttt{tberg@eng.ucsd.edu}}}

\begin{document}
\maketitle

\input{sections/abstract}
\input{sections/introduction}
\input{sections/related}
\input{sections/model}
\input{sections/experiments}
\input{sections/analysis}
\input{sections/conclusion}
\input{sections/limitations}
\input{sections/acknowledgements}

\bibliography{journal-full,custom}
\bibliographystyle{acl_natbib}

\newpage
\appendix

\input{sections/appendix}

\end{document}

%% file: math_commands.tex
\usepackage{amsmath,amsfonts,bm}

\def\eqref#1{equation~\ref{#1}}

\def\1{\bm{1}}

\DeclareMathAlphabet{\mathsfit}{\encodingdefault}{\sfdefault}{m}{sl}
\SetMathAlphabet{\mathsfit}{bold}{\encodingdefault}{\sfdefault}{bx}{n}

\newcommand{\E}{\mathbb{E}}



%% file: sections/abstract.tex
\begin{abstract}
Contrastive learning has been successfully used for retrieval of semantically aligned sentences, but it often requires large batch sizes and carefully engineered heuristics to work well. In this paper, we instead propose a generative model for learning multilingual text embeddings which can be used to retrieve or score sentence pairs. Our model operates on parallel data in $N$ languages and, through an approximation we introduce, efficiently encourages source separation in this multilingual setting, separating semantic information that is shared between translations from stylistic or language-specific variation. We show careful large-scale comparisons between contrastive and generation-based approaches for learning multilingual text embeddings, a comparison that has not been done to the best of our knowledge despite the popularity of these approaches. We evaluate this method on a suite of tasks including semantic similarity, bitext mining, and cross-lingual question retrieval---the last of which we introduce in this paper. Overall, our Variational Multilingual Source-Separation Transformer (\model) model outperforms both a strong contrastive and generative baseline on these tasks.\footnote{Code and Flax-based T5X model checkpoint available at \url{https://github.com/google-research/google-research/tree/master/vmsst}.}
\end{abstract}

%% file: sections/introduction.tex
\section{Introduction}

Contrastive learning is the dominant paradigm for learning text representations from parallel text~\cite{hermann-blunsom-2014-multilingual,singla-etal-2018-multi,guo-etal-2018-effective,wieting-etal-2019-simple,feng-etal-2022-language}. However, contrastive learning requires strong negative examples in the data and finding these negatives can be expensive in terms of compute or manual effort. In this paper, we propose a generative\footnote{We mean generative both in terms of text generation and as a statistical model of the joint probability distribution.} 
model for learning multilingual text embeddings which encourages source separation, separating semantic information that is shared between translations from stylistic or language-specific variation. We find that by filtering this variation into separate variables, performance of the remaining representations, that encode shared semantic information, increases across all downstream tasks.

Through an approximation that greatly reduces the memory footprint of our model, we scale our model and train on \N languages. We systematically compare our model, the \modelfull (\model) to strong contrastive and generative baselines on a suite of tasks including semantic similarity, bitext mining, and question retrieval, which we introduce for the cross-lingual setting, using the same training data and architecture. We show that our model outperforms these models and is also competitive with the state-of-the-art.

We analyze \model with careful ablations, showing the contribution of each aspect of the model to performance. We also show that even at large batch sizes, the advantage over contrastive learning remains, especially for large models. Furthermore, we also find the learned embedding space of our model to be smoother, making it less affected by the ``hubness problem''~\cite{radovanovic2010hubs,radovanovic2010existence} in representation learning, and more suitable for large-scale retrieval than the baseline methods. 

To the best of our knowledge, this is the first work to systematically compare generative and contrastive models for learning multilingual embeddings on a large parallel corpus containing many languages in a carefully controlled experimental setup---despite the popularity of these approaches~\cite{artetxe-schwenk-2019-massively,yang-etal-2020-multilingual}. We carry out these experiments with both pretrained and randomly initialized models. The comparison of objective functions is an important research question due to the large amounts of multilingual text available to train models and the many uses of these models in downstream tasks. To that end, another contribution of this paper is showing these comparisons and the surprising result that contrastive objectives do not provide the overall best accuracy on downstream tasks. Moreover, our generative \model increasingly outperforms the contrastive model when more layers are added and when training with larger batches and more training data, suggesting that as models continue to scale in the future, this performance gap may continue to increase further motivating the use of generative approaches for learning multilingual text embeddings.

%% file: sections/related.tex
\section{Related Work}

There has been a number of approaches proposed for learning bilingual and multilingual text embeddings. One popular approach is contrastive learning~\cite{hermann-blunsom-2014-multilingual,singla-etal-2018-multi,guo-etal-2018-effective,wieting-etal-2019-simple,feng-etal-2022-language} where translation pairs are positive examples and text from other pairs are used as negative examples. An alternative approach is to use a neural machine translation objective, where the representation from the hidden states of the encoder is used as the sentence embedding~\cite{espana2017empirical,schwenk-douze-2017-learning,artetxe-schwenk-2019-massively}. Other approaches include multi-task learning approaches which often use some type of contrastive learning of parallel text to align representations among languages~\cite{yang-etal-2020-multilingual,goswami-etal-2021-cross}, cross-lingual pretraining~\cite{chi-etal-2022-xlm}, and model distillation from a large pretrained multilingual model~\cite{reimers-gurevych-2020-making}.

An alternative approach that is more closely related to our work is generative models that separate the linguistic variation from the shared semantic information in translation pairs. \citet{wieting-etal-2020-bilingual} considered this for bitext, with each language having its own encoder and decoder parameters. This approach however does not scale, since it is not feasible to have thousands of encoders and decoders if one wants to model all of the more than 7,000 languages in the world.

%% file: sections/model.tex
\section{Model}

\begin{figure}[t]
    \centering
    \small
    \includegraphics[scale=0.24, trim = {0cm 0cm 0cm 0cm}]{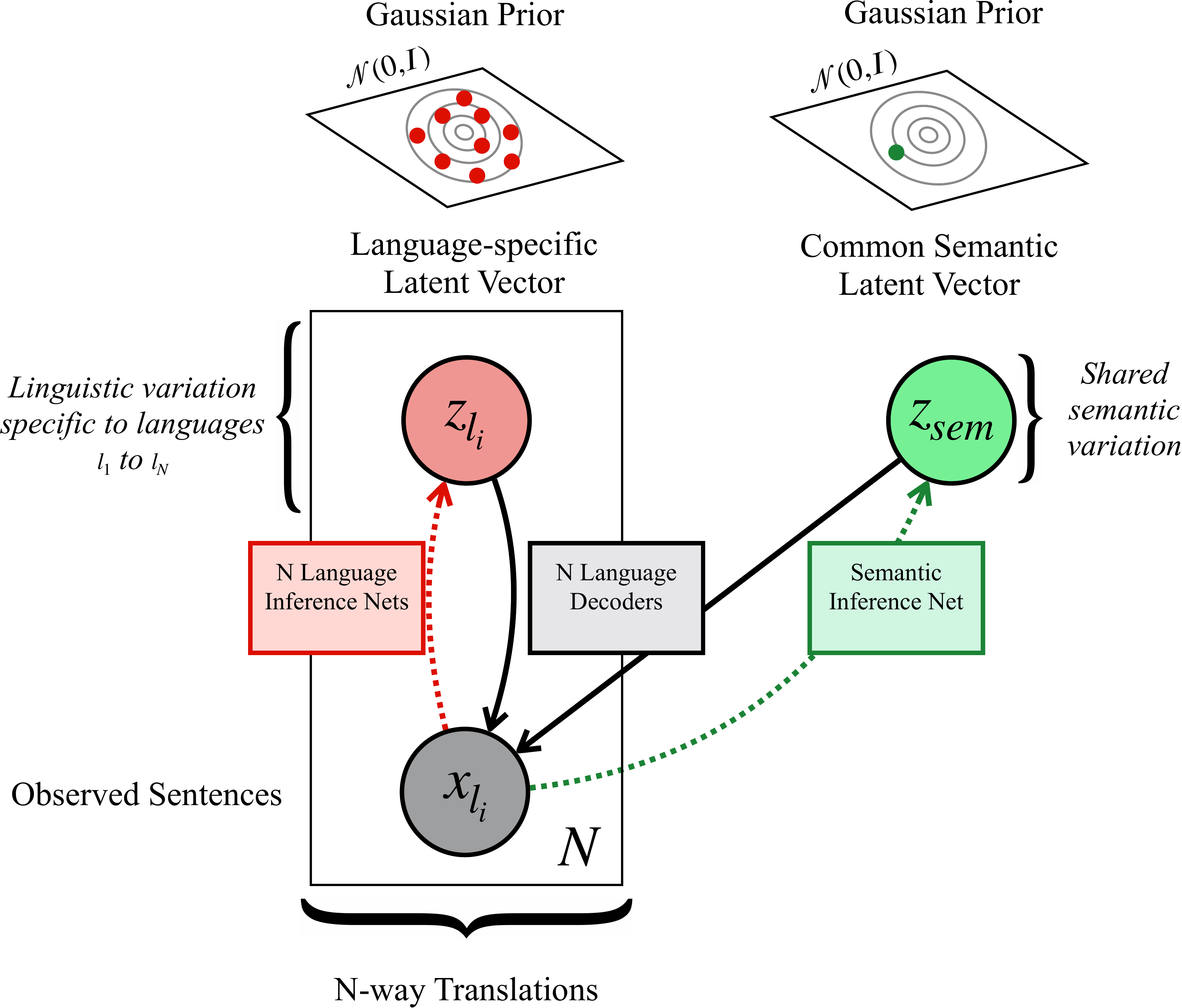}
    \caption{The generative process of our model. Latent variables, $z_{l_i}$, modeling the variation $x_{l_i}$ specifically due to language $l_i$, as well as a latent variable modeling the common semantics, $z_{sem}$, are drawn from a multivariate Gaussian prior. The observed translation in each language is then conditioned on its language-specific variable and $z_{sem}$. In practice, we approximate this model to make learning and inference tractable.
    }
    \label{fig:generative}
\end{figure}

\begin{figure*}
    \centering
    \small
    \includegraphics[scale=0.31]{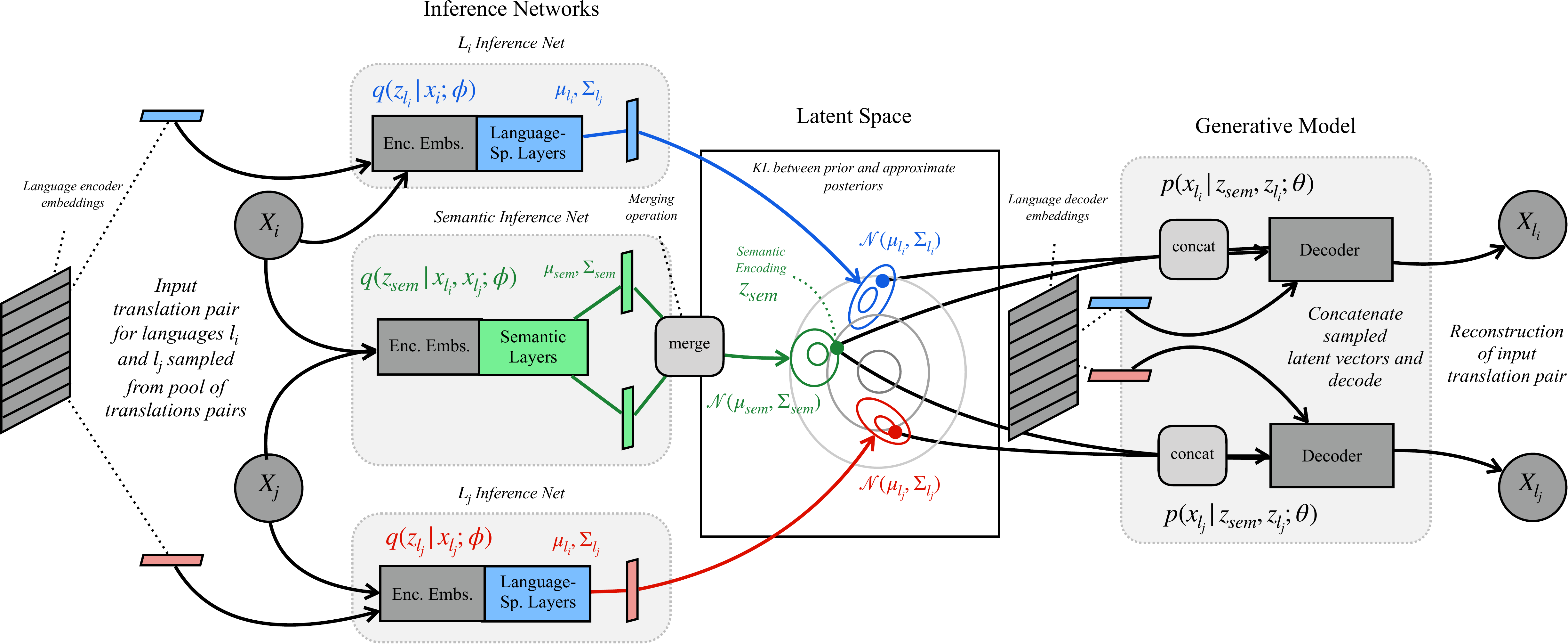}
    \caption{The computation graph for the variational lower bound used to train \model. The text for languages $l_i$ and $l_j$, with their respective language embeddings, are fed into the encoder acting as their inference networks. The text is also fed into the semantic inference network which is a separate encoder. The output of these networks are the language variables $z_{l_i}$ and $z_{l_j}$ and semantic variable $z_{sem}$. Each language-specific variable is then concatenated to $z_{sem}$ and used by a single shared decoder to reconstruct the input sentence pair. 
    }
    \label{fig:computationgraph}
\end{figure*}

The generative process of our underlying probabilistic model and the computation graph of our training objective procedure are depicted in Figure~\ref{fig:generative} and Figure~\ref{fig:computationgraph} respectively. In the generative story for \model, we first sample a semantic variable $z_{sem}$ for the sentence. Then for each of the $N$ languages, we sample a language-specific variable $z_{l_i}$. Each latent variable $z$ is sampled from a multivariate Gaussian prior $\mathcal{N}(0,I_k)$. These variables are then fed into a decoder that samples each of the $N$ sentences in the translation set. Each observed translation $x_{l_i}$, is sampled conditioned on $z_{sem}$ and its language variable $z_{l_i}$. Because $z_{sem}$ will be used to generate the sampled sentences in all languages, we expect that this variable will encode semantic, syntactic, or stylistic information that is shared in all of the translations. Conversely, the language variables $z_{l_i}$ will handle language-specific peculiarities or specific style differences that are not central to the meaning of the translation and are therefore not contained in many of the sentences. Concretely, the likelihood function of our model can be written for a single $N$-way tuple of translations $x = (x_1, ..., x_N)$:
\vspace{-1pt}
$$
p(x | z_{sem}, z_{l_1}, ..., z_{l_N}) = \prod_i^N p(x_i|z_{sem}, z_{l_i})
$$
\vspace{-3pt}

In the next section, we discuss how this separation of information is encouraged during learning.

\section{Learning and Inference} \label{sec:learning}

We would like to train our model on a set of parallel sentences $X$ consisting of $M$ examples in $N$ languages and a collection of latent variables $Z$. However, $N$-way parallel corpora are not available at the scale of bilingual text, and so we therefore approximate an $N$-way parallel corpus by sampling translation pairs from a large pool of pairs containing text in $N$ languages. Therefore in our model, $X=\{\langle x^1_{l_i}, x^1_{l_j} \rangle, \ldots, \langle x^M_{l_i}, x^M_{l_j} \rangle\}$ and $Z=(\langle z^1_{l_i}, z^1_{l_j}, z^1_{sem} \rangle, \ldots, \langle z^M_{l_i}, z^M_{l_j}, z^M_{sem} \rangle)$.

We aim to maximize the likelihood of the observed $X$ with respect to the parameters of the decoder $\theta$, marginalizing over the latent variables $Z$. We follow established procedures for this optimization problem from related latent variable models like variational autoencoders (VAEs; \citet{kingma2013auto}). Specifically, we optimize a variational lower bound on the log marginal likelihood, the evidence lower bound (ELBO). ELBO introduces a variational approximation $q(z_{sem},z_{l_i},z_{l_j}|x_{l_i}, x_{l_j};\phi)$ to the true posterior of the model. The $q$ distribution is parameterized by encoders or inference networks with parameters $\phi$. ELBO can be optimized by gradient ascent by using the reparameterization trick~\citep{kingma2013auto}, which allows for the expectation under $q$ to be approximated through sampling in a way that preserves backpropagation. The decoders and encoders are discussed in further detail in Section~\ref{sec:architecture}.

In contrast to variational autoencoders, which have only a single latent variable for each example, we have three in our model for each example. To encourage source separation, we make several independence assumptions for $q$ and factor it into three terms:

\begin{align*}
q(&z_{sem},z_{l_i},z_{l_j}|x_{l_i}, x_{l_j};\phi) = \\ &q(z_{sem} | x_{l_i}, x_{l_j};\phi) q(z_{l_i} | x_{l_i};\phi) q(z_{l_j} | x_{l_j};\phi)
\end{align*}

Lastly, we note that the ELBO contains a KL term that acts to regularize the latent variables. In our model, the KL term encourages $z_{sem}$, $z_{l_i}$, and $z_{l_j}$ to be close to a zero-centered Gaussian prior. The KL term thus encourages source separation, as encoding information shared by the translation pair in the shared variable results in only a single penalty from the KL loss, while encoding the information separately in the language-specific variables unnecessarily doubles the overall cost. In effect, we can view these language-specific latent variables as collecting information that cannot be captured in a common semantic space, separating it out from the variables collecting shared semantic information that we use for downstream tasks.

\paragraph{Objective Function.}
The overall objective function for \model consists of consists of two terms, the first being ELBO as described earlier:
\begin{multline*}
\text{ELBO} = \E_{q(Z_{S}, Z_{L}|X;\phi)} [\log p(X|Z_{S}, Z_{L};\theta)] -\\
       \text{KL}(q(Z_{S}, Z_{L}|X;\phi) || p(Z_{S};\theta)p(Z_{L};\theta))
\end{multline*}
where $Z_S$ is the collection of semantic variables, while $Z_L$ is the collection of language variables.

The second term, which we found necessary for strong performance, is the sum of $p(x_{l_i} | \mu_{sem_{l_j}})$ and  $p(x_{l_j} | \mu_{sem_{l_i}})$ which can be interpreted as samples from the mean of the posterior distribution using semantic variables generated from both input sentences. When training variational objectives, where the model ignores the latent variables and the learned posterior remains close to the prior.
Examples of other approaches to address these issues include: \cite{yang2017improved,kim2018semi,xu-durrett-2018-spherical,he2019lagging}. We weight the ELBO by $\lambda$ giving the total objective as:

\begin{align*}
\sum_{(x_{l_i}, x_{l_j}) \in X} \hspace{-2pt} p(x_{l_i} | \mu_{sem_{l_j}}) + p(x_{l_j} | \mu_{sem_{l_i}}) + \lambda\hspace{1pt}\text{\small ELBO}
\end{align*}

Therefore, our objective resembles translation with a weighted source-separation term. We show the effectiveness of this formulation compared to a pure translation objective in our experiments in Section~\ref{sec:experiments}.

\section{Architecture} \label{sec:architecture}

Our architecture is an encoder-decoder model, where the encoder produces a single representation that is fed into the decoder. Cross-attention between the encoder and decoder is not used, therefore the decoder has no full sequence visibility and more pressure is applied on the encoder to create a semantically meaningful representation. Specifically, we follow the approach of \citet{wieting-etal-2020-bilingual} which uses a Transformer~\cite{vaswani2017attention} encoder-decoder model, where the sentence embeddings are used in two places: at each layer of the decoder in place of cross-attention and in the computation of the logits.

\paragraph{Decoder Architecture.}
The decoder models $p(x_{l_i}|z_{sem}, z_{l_i};\theta)$ for each language $i$ (see right side of Figure~\ref{fig:computationgraph}). The inputs to the decoder are the language-specific variable $z_{l_i}$ and the semantic variable $z_{sem}$, which are concatenated and used to condition the decoder to generate the reconstruction of the observed text $x_{l_i}$. We use a single decoder for all languages.

\paragraph{Encoder Architecture.}
The encoders play an important role in the source separation as well as inference as detailed below. 

In order to motivate the separation of the linguistic and semantic information we split the encoder into two parts, only sharing the embedding table. We use one of these encoders to be the semantic inference network, which produces the semantic variable. The other encoder represents the $N$ language inference networks and produces the language variables for each language. These inference networks are shown on the left side of Figure~\ref{fig:computationgraph}. We mean-pool the hidden states followed by a linear projection to produce each variable from the encoders.

The semantic inference network, which models $q(z_{sem} | x_{l_i}, x_{l_j};\phi)$, is a multilingual encoder that encodes each language. For each translation pair, we alternate which of the two parallel sentences is fed into the semantic encoder within a batch for the ELBO term in the objective. Since the semantic encoder is meant to capture language agnostic semantic information, its outputs for a translation pair should be similar regardless of the language of the input sentence. We use the mean of the semantic encoder as the sentence representation for downstream tasks.

%% file: sections/experiments.tex
\section{Experiments} \label{sec:experiments}

\subsection{Constructing the Training Data}

We follow~\citet{artetxe-schwenk-2019-massively} in constructing our training data. However, since the exact data is not publicly available, we expect their may be small differences due to random sampling and different dataset versions. More specifically we sample our data from Europarl,\footnote{\url{http://opus.nlpl.eu/Europarl.php}} United Nations~\cite{rafalovitch-dale-2009-united},\footnote{\url{https://opus.nlpl.eu/UN.php}} OpenSubtitles2018~\cite{lison-etal-2018-opensubtitles2018},\footnote{\url{http://opus.nlpl.eu/OpenSubtitles.php}}  Global Voices,\footnote{\url{https://opus.nlpl.eu/GlobalVoices.php}} Tanzil,\footnote{\url{https://opus.nlpl.eu/Tanzil.php}} and Tatoeba v2021-07-22.\footnote{\url{https://opus.nlpl.eu/Tatoeba.php}}

We sample the same amount of data as was done in~\citet{artetxe-schwenk-2019-massively}, detailed in Appendix~\ref{sec:appendix-full-training-data}. 
The only deviation being that we take care to not include any Tatoeba test data in our training data. Our final corpus has nearly 216 million training examples, slightly less than 220 million reported in~\citet{artetxe-schwenk-2019-massively}. We use both English and Spanish as pivot languages, so each pair includes at least one English or Spanish sentence, and we use approximately the same amount of data for each language. We note that we only have training data for 92 languages instead of the 93 in~\citet{artetxe-schwenk-2019-massively} due to not having training data for Aymara (ay).

\subsection{Evaluation}

We evaluate on three tasks: semantic similarity, bitext mining and question retrieval. While the first two are commonly used to evaluate multilingual sentence embeddings, we introduce question retrieval in this paper. As can be seen by our results, we found question retrieval to be somewhat uncorrelated to either of the latter two. For each task, we use a collection of different datasets, detailed below.

\paragraph{Semantic Textual Similarity}
The goal of the semantic textual similarity tasks is to predict the degree to which sentences have the same meaning as measured by human judges. The evaluation metric is Pearson’s $r \times 100$ with the gold labels, which is convention for these tasks.

We make a distinction between two semantic similarity evaluations, English-only and cross-lingual. For the English-only evaluation, we follow~\citet{wieting-16-full} by averaging the yearly performance on 2012--2016 SemEval Semantic Textual Similarity (STS) shared tasks~\cite{agirre-etal-2012-semeval,agirre-etal-2013-sem,agirre-etal-2014-semeval,agirre-etal-2015-semeval,agirre-etal-2016-semeval}. More specifically, for each year of the competition, we average the Pearson's $r \times 100$ for each dataset in that year, and then finally average this result for each year of the competition. For the cross-lingual evaluation we use the cross-lingual STS tasks from SemEval 2017~\cite{cer-etal-2017-semeval}. This evaluation contains Arabic-Arabic, Arabic-English, Spanish-Spanish, Spanish-English, and Turkish-English STS datasets. These datasets were created by translating one or both pairs of an English STS pair into Arabic (ar), Spanish (es), or Turkish (tr). We average Pearson's $r \times 100$ for these datasets.

\paragraph{Bitext Mining}
For bitext mining, we use the Tatoeba dataset introduced in \citet{artetxe-schwenk-2019-massively} and the 2018 Building and Using Parallel Corpora (BUCC) shared bitext mining task~\cite{zweigenbaum2018overview}. 

The Tatoeba dataset consists of 100--1000 pairs of data aligned to English for 112 languages. The accuracy for Tatoeba can be computed in two ways, depending if English is the target language or source language. We compute accuracy using cosine similarity in both directions for all 112 languages (19 are unseen in the training data) and average this score for all languages. 

The goal of the BUCC task is to find the gold {\it aligned} parallel sentences given two corpora (one being very large) in two distinct languages. Languages are aligned with English and consist of German (de), French (fr), Russian (ru), and Chinese (zh). Typically, only about 2.5\% of the sentences are aligned. Following \citet{schwenk-2018-filtering}, we evaluate on the publicly available BUCC data. This involves scoring all pairs between the source target sentences and finding the optimal threshold that separates the data. Using the threshold, we can compute the precision, recall, and $F_1$ of the alignments. We report $F_1 \times 100$ in our results.

We compare two different approaches for finding the sentence alignments. In the first, BUCC (cosine), we compute the cosine similarity between the non-English source sentences and the English target sentences, selecting the highest scoring English sentence as the match. In the second, BUCC (margin), we follow \citet{artetxe-schwenk-2019-margin} and use a margin-based scoring approach, where the final score of a sentence pair is both a function of the score between the pair and the scores of each sentence with its nearest neighbors. To compute this margin score, we divide the cosine similarity for source sentence $s_i$ and target sentence $t_i$ by the sum of the scores of the four nearest neighbors of $s_i$ with the target sentences and the sum of the scores of the four nearest neighbors of $t_i$ with the source sentences.

Margin-based scoring is designed to alleviate the ``hubness problem''~\cite{radovanovic2010hubs,radovanovic2010existence} where the neighborhood around embeddings in a high-dimensional space, like in sentence embeddings, have many neighbors in common. These neighbors can displace the correct mapping in the ordering, hurting performance.

\paragraph{Question Retrieval}

For our question retrieval evaluation, we report the accuracy (R@1) on the test sets of Natural Questions (NQ)~\cite{kwiatkowski-etal-2019-natural} and the Multilingual Knowledge Questions and Answers (MKQA)~\cite{longpre-etal-2021-mkqa}. We use the the Probably Asked Questions dataset (PAQ)~\cite{lewis-etal-2021-paq} as a knowledge base from which we look up the nearest neighbor of each question in the NQ and MKQA test sets using cosine similarity. PAQ is a very large resource of 65 million automatically generated question-answer pairs. This is a zero-shot evaluation without any NQ supervised data.\footnote{This is opposed to the formulation in the original paper where a model based on BART Large~\cite{lewis-etal-2020-bart} was fine-tuned using a RAG-like objective~\cite{lewis2020retrieval} on the NQ training data in a model the authors call RePAQ. RePAQ, without using a reranker achieves an accuracy of 41.2 on NQ.}

\paragraph{Overall Score} We consolidate all of these evaluations into a score, as a way to get a sense of overall performance since different models favor different evaluations. While we are averaging different metrics (accuracy, Pearson's $r$, and $F_1$), we justify this as they do have the same scale,\footnote{Technically Pearson's $r$ can be negative, but this does not happen in our evaluations.} and a simple average gives a way for us to see overall performance. Our score is the average of six subtasks, two subtasks for each of semantic similarity, bitext mining, and question retrieval: English semantic similarity, cross-lingual semantic similarity, Tatoeba, BUCC (we average performance of the cosine and margin based scoring), NQ, and MKQA.

\subsection{Baselines} \label{sec:contrastive}
We compare \model against two strong baselines, which have been used extensively in the literature. 

The first baseline is \contrastive, where we use contrastive learning with the other sentences in the batch (``in-batch negative sampling'') as negative examples~\citep{sohn2016improved}. \contrastive is computed as the average of computing $p(s_i | t_i)$ and $p(t_i | s_i)$ for source sentence $s_i$ and target sentence $t_i$, and their respective representations $\mathbf{s_i}$ and $\mathbf{t_i}$ where the first term uses all the other targets as negatives and the second use all of the other source sentence as negatives. Specifically,

\begin{align*}
p(s_i | t_i) &= \exp (\mathbf{s_i} \cdot \mathbf{t_i}) ~/~ \sum_{j \in B }\exp \mathbf{s_i} \cdot \mathbf{t_j}\\
p(t_i | s_i) &= \exp (\mathbf{s_i} \cdot \mathbf{t_i}) ~/~ \sum_{j \in B }\exp \mathbf{t_i} \cdot \mathbf{s_j}\\
\text{loss} = - \frac{1}{2|\mathcal{B}|}&\sum_{(s_i, t_i) \in \mathcal{B}} \log p(s_i | t_i) + \log p(t_i | s_i)
\end{align*}

\noindent where $\mathcal{B}$ is a minibatch. This version of contrastive learning has been used in representation learning for retrieval (DPR,~\citealp{karpukhin-etal-2020-dense}), visual tasks (SimCLR,~\citealp{chen2020simple}) and image/text tasks (CLIP,~\citealp{pmlr-v139-radford21a}). There are other variations of this loss~\cite{qian2019softtriple}, and other contrastive losses like triplet loss~\cite{weston2010large} which has been used for learning text embeddings, but we leave a comparison of contrastive objectives for learning multilingual text embeddings for future work. 

The second baseline is \bitrans, where we use a translation objective to learn the representation~\citet{espana2017empirical,schwenk-douze-2017-learning,artetxe-schwenk-2019-massively}. 

We also explore an alternative to the \model, \mgtcontrastive, by incorporating a contrastive loss to use in a multitask setting. Again, we weight the contribution of the \model loss by $\lambda$. 

\begin{table*}[t!]
\small
\setlength{\tabcolsep}{3pt}
\begin{center}
\begin{tabular}{ lrrrr|rrr|rr|r } 
 \toprule
 \cmidrule{2-10} \vspace{-0.3cm} \\
 Model & \multicolumn{4}{c|}{Sem. Sim.} & \multicolumn{3}{c|}{Bitext Mining} & \multicolumn{2}{c|}{Quest. Retrieval} & Score  \\
 & Eng. & XL & XL (s.) & XL (d.) & Tatoeba & BUCC (c.) & BUCC (m.) & NQ & MKQA & \\
 \midrule
 \multicolumn{3}{l}{\textbf{Random Init. (6 Layer)}}\vspace{0.15cm}\\
 \contrastive & 65.5 & 66.8 & 73.3 & \bf 62.4 & 63.1 & 66.2 & 84.0 & 34.1 & 17.6 & 53.7\\
 \bitrans & 69.6 & 63.9 & 71.6 & 58.7 & 53.3 & 62.1 & 81.2 & \bf 37.4 & 19.2 & 52.5\\
 \mgtcontrastive & 65.7 & 66.3 & 73.0 & 61.9 & \bf 63.2 & 65.8 & 84.3 & 34.1 & 17.7 & 53.7\\
 \model & \bf 70.1 & \bf 67.4 & \bf 75.1 & 62.2 & 58.7 & \bf 73.7 & \bf 85.9 & 37.3 & \bf 20.1 & \bf 55.6\\
 \midrule
  \multicolumn{3}{l}{\textbf{Random Init. (24 Layer)}}\vspace{0.15cm}\\
 \contrastive & 64.4 & 64.6 & 71.6 & 60.0 & 62.7 & 64.3 & 83.7 & 32.8 & 16.0 & 52.4\\
 \bitrans & \bf 71.2 & 68.1 & 74.6 & 63.8 & 57.4 & 70.8 & 86.9 & 38.2 & 21.6 & 55.9\\
 \mgtcontrastive & 68.2 & 69.7 & 75.5 & 65.9 & \bf 64.8 & 58.5 & 84.1 & 36.9 & 18.9 & 55.0\\
 \model & 71.1 & \bf 71.7 & \bf 77.7 & \bf 67.7 & 61.4 & \bf 78.7 & \bf 89.0 & \bf 38.3 & \bf 22.3 & \bf 58.1\\
 \midrule
  \multicolumn{3}{l}{\textbf{Pretrained (24 Layer)}}\vspace{0.15cm}\\
 \contrastive & 73.3 & 74.7 & 76.0 & 73.9 & 85.1 & 74.3 & \bf 93.7 & 40.2 & 27.6 & 64.2\\
 \bitrans & 74.0 & 78.0 & 79.8 & 76.8 & 78.2 & 85.9 & 91.9 & \bf 40.9 & 29.6 & 64.9\\
 \mgtcontrastive & 73.4 & 75.4 & 76.7 & 74.6 & \bf 85.4 & 74.6 & \bf 93.7 & 40.3 & 27.9 & 64.4\\
 \model & \bf 74.6 & \bf 79.1 & \bf 81.5 & \bf 77.5 & 81.1 & \bf 87.8 & 92.5 & 40.8 & \bf 29.9 & \bf 65.9\\
\bottomrule
\end{tabular}
\end{center}
\caption{Experimental results for \model and \mgtcontrastive and our baselines \contrastive and \bitrans. We evaluate on semantic similarity, bitext mining, and question retrieval. For semantic similarity we separate the evaluations into English-only, cross-lingual, cross-lingual but with the same language (XL (s.) ar-ar and es-es) and cross-lingual using different languages (XL (d.), ar-en, es-en, and tr-en). Results are reported as the average Pearson's $r \times 100$ across datasets. For bitext mining we evaluate on Tatoeba and BUCC, with BUCC split between using cosine similarity or using a margin approach~\cite{artetxe-schwenk-2019-margin}. Results are reported as accuracy $\times 100$ for Tatoeba and $F_1 \times 100$ for BUCC. For question retrieval, we evaluate retrieval accuracy $\times 100$ using PAQ as a question knowledge base on the NQ and MKQA datasets. Finally, we compute a score to summarize quality over these evaluations.
}
\label{tab:experimental-results}
\end{table*}

\subsection{Experimental Settings}

We explore three different settings for each of four objective functions we consider. We use the Transformer architecture for all settings. Specifically, we explore a 6 layer encoder-decoder model, a 24 layer encoder-decoder model, and a 24 layer encoder-decoder initialized with the Multilingual T5 (mT5) Large~\cite{xue-etal-2021-mt5}. We set the dimension of the embeddings and hidden states for the encoders and decoders to 1024. The mT5 Large model inherently has embedding and hidden state dimensions of 1024. For all models, we use the mT5 vocabulary, which is derived from \texttt{sentencepiece}~\cite{kudo-richardson-2018-sentencepiece}. The vocabulary consists of 250,000 tokens and was learned from multilingual variant of the C4 dataset called mC4 which includes 101 languages.

For optimization, we use Adafactor~\citep{shazeer2018adafactor}. We use the same learning rate schedule as~\citet{vaswani2017attention}, i.e., the learning rate increases linearly for 4,000 steps, after which it is decayed proportionally to the inverse square root of the number of steps. We set the peak learning rate to be 0.001, and we train our models for 100,000 steps total. We use a batch size of 2048 and set the maximum sequence length of our model to 32 for all experiments.

We use a dropout rate of 0.1 for \contrastive models and no dropout for \bitrans, \mgtcontrastive (with the exception of the randomly initialized 24 layer model which used 0.1), and \model. For \model, we anneal the KL term so that it increased linearly for 1,000,000 updates.

For \model, we set $\lambda$, the weight on the \model ELBO loss term, to be 0.025 for the pretrained models, and 0.1 when training from randomly initialized parameters. For \mgtcontrastive, we set it to .0005 for the pretrained and 6 layer settings and 0.001 for the randomly initialized 24 layer setting.

\subsection{Results}

The results of our experiments are shown in Table~\ref{tab:experimental-results}. Overall, \model has the best performance for all three experimental settings and the best performance on each task on average, with the exception of Tatoeba. In fact, for NQ question retrieval with a pretrained model, it performs nearly to that of the model trained specifically for this task on NQ data from~\citet{lewis-etal-2021-paq} which has an accuracy of 41.2. \model and \bitrans are especially strong when using more layers, which is not the case for \contrastive which declines in performance when moving from 6 to 24 layers. In fact at 24 layers, \bitrans performs better on average than \contrastive. Perhaps for even larger models, the gap between contrastive and generative models will increase. We also see that \contrastive seems to benefit more from pretraining than \model and \bitrans, which could possibly be due to \model re-purposing and adding additional randomly initialized parameters to the decoder. Perhaps different pretraining strategies using this modified decoder would resolve these differences. We also see that \mgtcontrastive has negligible improvement over \contrastive which was unexpected ---that is, a traditional contrastive loss does not improve further on top of generative loss of \model. We leave the exploration of different strategies of combining these approaches to future work.

It is also interesting to observe the stark performance difference for different tasks. Bitext mining tasks like Tatoeba, and BUCC (m.) for the pretrained 24 layer model, favor \contrastive, while semantic similarity, BUCC (c.) and question retrieval favor \model, suggesting some fundamental difference in these tasks favoring \contrastive. An examination of the Tatoeba and BUCC data shows that there are paraphrases in the test set, but accounting for these does not seem to meaningfully explain this performance difference.

Lastly, we see that \model outperforms \contrastive on the BUCC task with cosine similarity, though the results between the two models are closer when using margin. This suggests that the ``hubness problem''~\cite{radovanovic2010hubs,radovanovic2010existence} where the neighborhood around embeddings in a high-dimensional spaces have many neighbors in common, is less of an issue when learning embeddings with \model. This smoother embedding space may also contribute to the stronger results \model has on the question retrieval tasks.

\subsection{Comparison to Related Work}

Prior work on learning multilingual embeddings has explored a variety of models utilizing different strategies and using difference source and types of training data.
However, comparing approaches is difficult as they differ in many factors that are crucial to performance: training data, model size, architecture, vocabulary, training time, and evaluation datasets. Complicating matters further, even the metric used in evaluation for the same dataset, the distance measure used between embeddings for the same dataset, and the specific subsets of the evaluation datasets used can be different.

The main goal of this paper is to compare contrastive and generative losses systematically and uniformly, on the same data, metrics and underlying architecture. However, we also emphasize that the best systems we compare are competitive with the current state-of-the-art. Hence, in this section we compare \model to published results of other models on semantic similarity and the Tatoeba and BUCC bitext mining tasks. We primarily compare against five models which have the strongest multilingual results in the literature: mUSE~\cite{yang-etal-2020-multilingual}, LASER~\cite{artetxe-schwenk-2019-massively}, XLM-R (NLI/STS-B) and XLM (Para.)~\cite{reimers-gurevych-2020-making}, and LaBSE~\cite{feng-etal-2022-language}.

For semantic similarity, we include Spearman's $\rho$ in order to compare to work that solely uses this correlation metric. We use cosine as the similarity measure for all models in these evaluations.\footnote{Note that mUSE and LaBSE report results using the angle as the metric instead of its cosine for semantic similarity tasks, but as they do not evaluate on these specific datasets, we include the results from~\citet{reimers-gurevych-2020-making} for comparison which uses cosine similarity.} The results are shown in Table~\ref{tab:lit-comparison-sim}.

\begin{table}[ht]
\small
\begin{center}
\setlength{\tabcolsep}{3pt}
\begin{tabular}{ lrrr } 
\toprule
 Model & XL & XL (s.) & XL (d.)\\
 \midrule
 mUSE & 79.5 & 81.7 & 78.1\\
 LASER & 69.0 & 74.3 & 65.5\\
 XLM-R (NLI/STS-B) & 79.0 & 81.7 & 77.2\\
 XLM-R (Para.) & \bf 82.4 & \bf 82.9 & \bf 82.1\\
 LaBSE & 72.4 & 74.9 & 70.7\\
 \midrule
 \model & 79.4 & 81.9 & 77.7\\
\bottomrule
\end{tabular}
\end{center}
\caption{Comparisons to related work on cross-lingual semantic similarity. Results are reported in Spearman's $\rho \times 100$. XL contains all 5 datasets, where XL (s.) contains only those where the languages are the same (ar-ar, es-es), and XL (d.) contains those datasets where the languages are different (ar-en, es-en, and tr-en). Note that models in this table are not trained on the same data; for instance LaBSE was trained on substantially more parallel data and XLM-R (Para.) was trained using a large English paraphrase corpus in addition to parallel data.}
\label{tab:lit-comparison-sim}
\end{table}

For Tatoeba, we compare to methods that have evaluated on all 112 languages, which excludes mUSE as it was only trained on 16 language pairs. The results are shown in Table~\ref{tab:lit-comparison-tat}. Baselines results are taken from~\citet{reimers-gurevych-2020-making}.

\begin{table}[h!]
\small
\begin{center}
\setlength{\tabcolsep}{3pt}
\begin{tabular}{ lr } 
\toprule
 Model & Tatoeba\\
 \midrule
 LASER & 65.5 \\
 XLM-R (Para.) & 67.1 \\
 LaBSE & \bf 83.7 \\
 \midrule
 \model & 81.1 \\
\bottomrule
\end{tabular}
\end{center}
\caption{Comparisons to related work on Tatoeba. Results are reported as accuracy $\times 100$, averaging the xx->en and en->xx directions. Note that models in this table are not trained on the same data; for instance LaBSE was trained on substantially more parallel data and XLM-R (Para.) was trained using a large English paraphrase corpus in addition to parallel data.}
\label{tab:lit-comparison-tat}
\end{table}

\begin{table*}[th!]
\small
\setlength{\tabcolsep}{3pt}
\begin{center}
\begin{tabular}{ lrrrr|rrr|rr|r } 
 \toprule
 \cmidrule{2-10} \vspace{-0.3cm} \\
 Model & \multicolumn{4}{c|}{Sem. Sim.} & \multicolumn{3}{c|}{Bitext Mining} & \multicolumn{2}{c|}{Quest. Retrieval} & Score  \\
 & Eng. & XL & XL (s.) & XL (d.) & Tatoeba & BUCC (c.) & BUCC (m.) & NQ & MKQA & \\
 \midrule
  \multicolumn{3}{l}{\textbf{Random Init. (24 Layer)}}\vspace{0.15cm}\\
 \model & 71.1 & \bf 71.7 & \bf 77.7 & \bf 67.7 & 61.4 & 78.7 & 89.0 & 38.3 & 22.3 & 58.1\\
 \midrule
 \model (fact.) & 67.3 & 69.9 & 76.3 & 65.7 & \bf 63.0 & 77.9 & \bf 90.4 & 37.3 & 21.5 & 57.2\\
 \model (4 enc.) & \bf 71.2 & 70.2 & 76.6 & 66.0 & 60.8 & 77.7 & 88.5 & \bf 38.4 & 22.0 & 57.6\\
 \model (12L dec.) & 71.1 & 70.9 & 77.4 & 66.7 & 61.2 & 78.4 & 88.8 & 38.0 & 22.2 & 57.8\\
 \model (1L dec.) & 71.0 & 71.2 & 77.0 & 67.4 & \bf 63.0 & \bf 79.4 & 89.1 & 38.7 & \bf 22.8 & \bf 58.5\\
 \midrule
 \model (no KL) & 70.7 & 68.7 & 76.2 & 63.7 & 56.9 & 70.8 & 86.6 & 37.8 & 21.5 & 55.7\\
 \model (1 enc.) & 70.6 & 69.4 & 76.7 & 64.6 & 60.0 & 77.0 & 87.8 & \bf 38.4 & 21.4 & 57.0\\
 \model (no enc. l.e.) & \bf 71.2 & 69.8 & 76.1 & 65.5 & 61.2 & 78.7 & 88.9 & 38.2 & 22.0 & 57.7\\
 \model (no dec. l.e.) & 70.8 & 70.7 & 76.7 & 66.7 & 60.9 & 77.4 & 88.6 & 38.3 & 21.8 & 57.6\\
 \bottomrule
\end{tabular}
\end{center}
\caption{Ablations of \model. We investigate ablations involving factorization of the decoder projection layer (fact.), using 4 language encoders instead of 1 (4 enc.), using 12 layer (12L dec.) and 1 layer (1L dec.) decoders, using no KL term (no KL), using only a single encoder for both language and semantic variables (1 enc.), and using no encoder language embeddings (no enc. l.e.) or no decoder language embeddings (no dec. l.e.).}
\label{tab:ablations}
\end{table*}

For BUCC, we include results on the training sets using the margin retrieval methods from \citet{artetxe-schwenk-2019-massively}. 
The results are shown in Table~\ref{tab:lit-comparison-bucc}. Baselines results are taken from~\citet{artetxe-schwenk-2019-massively,reimers-gurevych-2020-making}.

\begin{table}[h!]
\small
\begin{center}
\setlength{\tabcolsep}{2pt}
\begin{tabular}{ lrrrrr } 
\toprule
 Model & de-en & fr-en & ru-en & zh-en & Avg.\\
 \midrule
 mUSE & 88.5 & 86.3 & 89.1 & 86.9 & 87.7\\
 LASER & 95.4 & 92.4 & 92.3 & 91.2 & 92.8\\
 XLM-R (NLI/STS-B) & 86.8 & 84.4 & 86.3 & 85.1 & 85.7\\
 XLM-R (Para.) & 90.8 & 87.1 & 88.6 & 87.8 & 88.6\\
 LaBSE & \bf 95.9 & \bf 92.5 & \bf 92.4 & \bf 93.0 & \bf 93.5\\
 \midrule
 \model & 94.3 & 91.0 & 91.8 & 92.8 & 92.5\\ 
\bottomrule
\end{tabular}
\end{center}
\caption{Comparisons to related work on BUCC in accuracy $\times 100$ using the margin approach from ~\citet{artetxe-schwenk-2019-margin}. Note that models in this table are not trained on the same data; for instance LaBSE was trained on substantially more parallel data and XLM-R (Para.) was trained using a large English paraphrase corpus in addition to parallel data.}
\label{tab:lit-comparison-bucc}
\end{table}

While \model does not have the best performance relative to models from the literature on any single task, it does have the best overall performance if one averages the results for each task.\footnote{The average performance for \model is 84.3, versus 82.6 for LaBSE, and 79.3 for XLM-R (Para.)} While these models share much in common, namely using parallel text and some type of pre-training or pretrained model, there are differences in the exact data and models used, among other confounding variables. For instance, LaBSE used training data consisting of six billion parallel pairs across languages and was also trained on monolingual text using a masked language modelling objective. XLM-R (Para.) makes use of a 50 million example paraphrase corpus for distillation. In contrast, our setup most closely follows LASER, using an approximation of the 220M example parallel data used to train their model.

%% file: sections/analysis.tex
\section{Analysis}

\subsection{Model Ablations}

\begin{table*}[th!]
\small
\setlength{\tabcolsep}{2.5pt}
\begin{center}
\begin{tabular}{ lrrrrrr|rrr|rrrr|r } 
 \toprule
 \cmidrule{2-10} \vspace{-0.3cm} \\
 Model & \multicolumn{6}{c|}{Sem. Sim.} & \multicolumn{3}{c|}{Bitext Mining} & \multicolumn{4}{c|}{Quest. Retrieval} & Score  \\
 & Eng. & \multicolumn{5}{c|}{XL} & \multicolumn{3}{c|}{Tatoeba} & NQ & \multicolumn{3}{c|}{MKQA} & \\
 - & - & ar-en & ar-ar & es-en & es-es & tr-en & ar & es & tr & - & ar & es & tr &  \\
 \midrule
 \multicolumn{3}{l}{\textbf{Random Init. (6 Layer) - ar, en, es, tr}}\vspace{0.15cm}\\
 \contrastive & 68.6 & 81.7 & \bf 68.5 & \bf 68.3 & \bf 64.8 & \bf 69.7 & 97.9 & \bf 88.2 & \bf 98.1 & 36.4 & 24.6 & 13.0 & 22.6 & 58.1\\
 \bitrans & 69.0 & 82.4 & 63.1 & 57.8 & 58.7 & 67.3 & 97.6 & 84.3 & 96.4 & 37.8 & 25.4 & 12.2 & 21.1 & 57.0\\
 \model & \bf 70.6 & \bf 83.1 & 65.9 & 63.1 & 62.1 & 68.8 & 97.9 & 84.9 & 97.2 & 38.1 & \bf 27.2 & 14.4 & 24.1 & 58.5\\
 \model (full enc., full dec.) & 70.3 & 82.3 & 65.6 & 60.0 & 62.8 & 67.9 & \bf 98.4 & 87.9 & 97.8 & \bf 39.0 & \bf 27.2 & \bf 14.7 & \bf 24.2 & \bf 58.7\\
\bottomrule
\end{tabular}
\end{center}
\caption{Comparison of \model with a variation that has no parameter sharing, \model (full enc., full dec.). We experiment on 4 languages, so we have 5 encoders and 4 decoders.
}
\label{tab:analysis=namgt-results}
\end{table*}

In this section, we investigate different ablations of \model. The ablations are shown in Table~\ref{tab:ablations}. We start from the 24 layer randomly initialized \model, and change it to see how certain hyperparameters and model choices affect performance.

Our first experiment, \model (fact.) investigates what happens if we simply factor the final projection layer of the decoder. This can save a lot of memory in the model, as that projection layer is $3 \times d \times V$ where $d$ is the hidden state dimension size and $V$ is the size of the vocabulary.\footnote{We multiply by 3 because we have three embeddings, the hidden state, the language-specific vector, and the semantic vector.} If we factor the projection layer, we can reduce the space to $d \times V + 3d \times d$. In practice, this saves about 509 million parameters for our 24 layer models. However from the first row in Table~\ref{tab:ablations}, we see that this small change has a significant effect on performance, weakening results on semantic similarity and question retrieval tasks and strengthening results on bitext mining tasks.

In our second ablation, \model (4 enc.), we spread the model capacity of the language-specific encoder to 4 encoders, instead of the single encoder in our previous experiments. We allocate the languages randomly to the different encoders. We find that this doesn't improve results, perhaps because the 24 layer model has sufficient capacity to model all of the languages in one shared encoder. We could allocate languages to encoders based on language families, and perhaps this could fare better, but we leave that for future work.

Prior work~\cite{wieting-etal-2020-bilingual} shows that, a decoder that is weaker (i.e. less layers) can lead to stronger embeddings. This effect is presumably because there is more pressure on the sentence embedding to fully and clearly capture the semantics since it cannot rely on a strong decoder to fill in gaps. We found that that using a weaker single layer decoder (1L dec.), does indeed seem to improve performance. We also tried a 12 layer ablation (12L dec.), but that seemed to not have a significant improvement in the results.

The last four ablations investigate different modelling choices. In the first we eliminate the KL term (no KL), which has the most significant effect on performance, especially on cross-lingual tasks. In the second ablation, we use a single encoder instead of the twin encoders (1 enc.), one for semantic embeddings and one for language embeddings, we find that this has a modest overall effect on performance. Lastly, we eliminate the language embeddings. First we remove the language embedding inputs to the decoder (no enc. l.e.), then we experiment by removing the input language embeddings to the language-specific encoder (no dec. l.e.). We find these language embeddings have a smaller than expected impact on performance, perhaps because the large capacity of the decoder can ascertain the language being input or decoded.

\subsection{Testing the Parameter Sharing in \model}

Parameter sharing was needed in order efficiently perform source separation on $N$ languages. Specifically we collapsed the language encoders into a single encoder and we collapsed  the decoders into a single decoder. The \model approximates having $N$ language encoders by using an input embedding to indicate the language being considered. The same strategy is applied with the decoders as well, with the first input token to the decoder indicating the language to be generated.

In this section, we investigate what effect this parameter sharing has on \model by using $N$ encoders and decoders (full enc, full dec.). We experiment with 6 layer Transformer encoders and 4 languages Spanish, English, Arabic, and Turkish in order to keep the experiments tractable as in this setting we have 5 encoders and 4 decoders. The results are shown in Table~\ref{tab:analysis=namgt-results}.

The results indicate that the approximation appears to hold, as \model is much closer to the full model than \bitrans, which is an ablation of \model without the source separation. However, there is still a gap between the full encoder/decoder of \model and \model. We hypothesize however, that as the number of layers of the model increases, this performance gap also shrinks. The extra capacity of these layers will allow for the model to separate language-specific variations without having separate parameters for each language. Evidence for this hypothesis is in Table~\ref{tab:ablations} where having the language variation shared amongst 4 encoders instead of 1 actually appears to weaken performance overall.

\begin{table*}[th!]
\small
\setlength{\tabcolsep}{3pt}
\begin{center}
\begin{tabular}{ llrrrr|rrr|rr|r } 
 \toprule
 \cmidrule{2-10} \vspace{-0.3cm} \\
 Model & B. Size & \multicolumn{4}{c|}{Sem. Sim.} & \multicolumn{3}{c|}{Bitext Mining} & \multicolumn{2}{c|}{Quest. Retrieval} & Score  \\
 & & Eng. & XL & XL (s.) & XL (d.) & Tatoeba & BUCC (c.) & BUCC (m.) & NQ & MKQA & \\
 \midrule
 \multicolumn{3}{l}{\textbf{Random Init. (6 Layer)}}\vspace{0.15cm}\\
 \multirow{3}*{\contrastive} & 2048 & 65.5 & 66.8 & 73.3 & 62.4 & 63.1 & 64.7 & 82.9 & 34.0 & 17.6 & 53.5\\
 & 4096 & 67.5 & 69.3 & 75.4 & 65.3 & 66.0 & 71.5 & 87.0 & 35.3 & 19.2 & 56.1\\
 & 8192 & 69.4 & \bf 71.6 & \bf 76.8 & \bf 68.1 & \bf 68.6 & 76.2 & \bf 89.4 & 36.4 & 20.9 & \bf 58.3\\
 \midrule
 \multirow{3}*{\model} & 2048 & 70.1 & 67.4 & 75.1 & 62.2 & 58.7 & 72.6 & 84.7 & 37.2 & 20.2 & 55.4\\
 & 4096 & 70.2 & 67.4 & 75.3 & 62.1 & 58.5 & 73.1 & 86.0 & 38.2 & 20.3 & 55.7\\
 & 8192 & \bf 71.4 & 70.9 & 76.6 & 67.1 & 61.8 & \bf 77.9 & 88.0 & \bf 39.0 & \bf 22.4 & 58.1\\
 \midrule
  \multicolumn{3}{l}{\textbf{Random Init. (24 Layer)}}\vspace{0.15cm}\\
 \multirow{3}*{\contrastive} & 2048 & 64.4 & 64.6 & 71.6 & 60.0 & 62.7 & 62.8 & 82.5 & 32.8 & 16.0 & 52.2\\
 & 4096 & 66.6 & 68.6 & 75.1 & 64.3 & 65.7 & 70.9 & 86.8 & 34.7 & 18.1 & 55.4\\
 & 8192 & 68.0 & 70.2 & 76.2 & 66.2 & \bf 67.7 & 74.2 & 88.3 & 35.2 & 19.4 & 57.0\\
 \midrule
 \multirow{3}*{\model} & 2048 & 71.1 & 71.7 & 77.7 & 67.7 & 61.4 & 78.4 & 87.8 & 38.3 & 22.3 & 58.0\\
 & 4096 & 72.0 & 72.1 & 77.7 & 68.3 & 62.9 & 81.0 & 89.7 & 38.7 & 23.5 & 59.1\\
 & 8192 & \bf 72.7 & \bf 74.1 & \bf 79.0 & \bf 70.8 & 64.1 & \bf 82.0 & \bf 90.2 & \bf 39.0 & \bf 24.3 & \bf 60.1\\
\bottomrule
\end{tabular}
\end{center}
\caption{Comparison of \contrastive and \model using different batch sizes during training.}
\label{tab:batch-size}
\end{table*}

\subsection{Zero-Shot Bitext Mining}

\begin{table}[h!]
\small
\begin{center}
\begin{tabular}{ lrrr } 
\toprule
 Model & Tat. (seen) & Tat. (unseen) & $\Delta$\\
 \midrule
 \multicolumn{3}{l}{\textbf{Random Init. (6 Layer)}}\vspace{0.15cm}\\
 \bitrans & 59.3 & 24.0 & -\\
 \model & \bf 64.4 & \bf 30.6 & 1.5\\
 \midrule
  \multicolumn{3}{l}{\textbf{Random Init. (24 Layer)}}\vspace{0.15cm}\\
 \bitrans & 82.6 & 56.5 & -\\
 \model & \bf 84.9 & \bf 62.2 & 3.4\\
 \midrule
  \multicolumn{3}{l}{\textbf{Pretrained (24 Layer)}}\vspace{0.15cm}\\
 \bitrans & 63.7 & 26.5 & -\\
 \model & \bf 67.3 & \bf 32.6 & 2.5\\
\bottomrule
\end{tabular}
\end{center}
\caption{Results on languages seen during training (seen) and languages that were not seen during training (unseen) on the Tatoeba dataset.}
\label{tab:zero-shot}
\end{table}

The Tatoeba dataset contains parallel sentence pairs of English with 112 languages. Our model is trained using 93 of these languages, and therefore there are 19 languages we can use for a zero-shot evaluation of bitext mining. Table~\ref{tab:zero-shot} summarizes the results of this zero-shot evaluation for the two generation objectives, \bitrans and \model considered in this paper. The results are shown in Table~\ref{tab:zero-shot}. We also compute $\Delta$ which is the difference between the performance gap of \model and \bitrans on the seen and unseen languages. From the results, we see that \model does even better than \bitrans on unseen languages than unseen languages. Since \bitrans can be seen as an ablation of \model, i.e. \model without the source-separation loss, we see that the source-separation loss especially helps with generalization to new languages.

\subsection{Effects of Batch Size}

Lastly, we investigate how \model compares to \contrastive as batch size increases. It is common knowledge that contrastive models learn better representations when given harder negative examples. Since we are using in-batch negatives in our contrastive baseline, the increased batch size increases the chances of encountering harder negative examples and will generally increase performance up to the point where the negatives become false. Furthermore. bigger batch sizes are known to also improve results in models using the Transformer architecture, presumably due to less noisy gradients, which would improve the results of both \contrastive and \model. It is important to note that using bigger batch sizes, means seeing more examples (100,000 steps at a batch size of 2048 is about 1 pass through the data). However, parallel data is so numerous that training to convergence on the available data is not very practical. Therefore, these experiments do not separate out the gains from using a bigger batch size versus seeing more training data, but we argue that is not an important distinction to make due to the sheer amount (billions of pairs) of parallel data available.

We experiment with batch sizes of 4096 and 8192, double and quadruple the 2048 used in all experiments up to this point, for both the 6 layer and 24 layer randomly initialized versions of \contrastive and \model. All models are trained again for 100,000 steps. The results are shown in Table~\ref{tab:batch-size}.

From the results, we see that for the 6 layer model, increasing the batch size equalizes \model and \contrastive overall, however each performs better at different tasks. \contrastive has better performance on Tatoeba, XL semantic similarity, and BUCC with margin~\cite{artetxe-schwenk-2019-margin}, where \model has better performance on English semantic similarity, BUCC with cosine similarity, and the retrieval tasks. For the 24 layer variations, \model is better at every task, with the exception of Tatoeba, and has the highest overall score of any model in the table. The 24 layer \contrastive variation does not perform as well as the 6 layer version at any batch size, in contrast to \model where the 24 layer model always outperforms the 6 layer variation.

%% file: sections/conclusion.tex
\section{Conclusion}

We present \model, a generative massively multilingual text embedding model trained to separate semantic information from language-specific information. \model also outperforms strong contrastive and generative baselines on a variety of tasks. There are several avenues for future work including alternative pretraining objectives that better fit the use case of the decoder, explore incorporating monolingual data into the generative objective, investigate synergy between \model and contrastive methods as they seem to specialize in different tasks, and lastly scale up to bigger models, more data, and languages to further investigate \model versus contrastive methods.

%% file: sections/limitations.tex
\section*{Limitations}

Some of our experiments, specifically those in the ablations with large batch sizes, required significant computational resources. We trained these models on Google Cloud TPUv3 Pod slice with 128 chips for a few days. This experiment is important, as otherwise there would be questions on how the models compare at large batch sizes where contrastive models are known to work better. Due to training costs and in the interest of open research, we will open source our code and model checkpoints for the community to use and build upon.

Secondly, \model and \bitrans require decoding which which means they need more memory for the decoder and are slower during training. However one advantage of these models is that they can be trained with gradient checkpointing greatly reducing their memory requirements, which cannot be used for the contrastive models as that would reduce the effective batch size for finding negative examples. Moreover, during inference, there is no difference in the memory or speed requirements in \contrastive, \bitrans, or \model as only a single encoder is used in inference and there is no decoding.

%% file: sections/acknowledgements.tex
\section*{Acknowledgements}

We are grateful to Livio Baldini-Soares, Wenhu Chen, Zhuyun Dai, Tom Kwiatkowski, Jianmo Ni, Slav Petrov, Jason Riesa, and Pat Verga for useful discussions during the course of the project.

%% file: sections/appendix.tex
\cleardoublepage
\section*{Appendices accompanying ``\titlestr''}

\section{Full Experimental Results} \label{sec:appendix-full-results}

We include full results for our models using the pre-trained mT5 large checkpoint. We evaluate on English semantic similarity, Cross-lingual semantic similarity, question retrieval, and bitext mining.

\subsection{Semantic Similarity}

\begin{table*}[htp]
\begin{center}
\small
\begin{tabular}{ lrrrrr } 
\toprule
Model & \multicolumn{5}{c}{English Semantic Similarity} \\
\cmidrule{2-6} \vspace{-0.3cm} \\
& 2012 & 2013 & 2014 & 2015 & 2016 \\
\midrule
\contrastive & 69.7 & 61.1 & 76.4 & 81.4 & 77.7\\
\bitrans & 69.1 & 63.6 & 76.4 & 81.0 & 79.9\\
\mgtcontrastive & 70.2 & 61.6 & \bf 76.5 & 81.3 & 77.5\\
\model & \bf 70.5 & \bf 64.3 & \bf 76.5 & \bf 81.6 & \bf 80.1\\
\midrule
\contrastive & 68.0 & 74.9 & 69.1 & 79.9 & 76.9\\
\bitrans & 70.7 & \bf 77.9 & 72.2 & 81.8 & \bf 79.7\\
\mgtcontrastive & 68.4 & 75.1 & 69.2 & 80.2 & 76.8\\
\model & \bf 72.7 & \bf 77.9 & \bf 72.7 & \bf 82.1 & 79.2\\
\bottomrule
\end{tabular}
\end{center}
\caption{Full results on English STS. In the first part of the table, we show results, measured in Pearson's $r \times 100$, for each year of the STS tasks 2012-2016 as well as the average performance across all years. In the second part, we evaluate based on the Spearman's $\rho \times 100$ of the concatenation of the datasets of each year with the 2013 SMT dataset removed following~\cite{reimers-gurevych-2019-sentence}.}
\label{tab:english-sts-full-results}
\end{table*}

\begin{table*}[htp]
\begin{center}
\small
\begin{tabular}{ lrr|rr|rr|rr|rr } 
\toprule
Model & \multicolumn{10}{c}{Cross-Lingual Semantic Similarity} \\
\cmidrule{2-11} \\
& \multicolumn{2}{c}{ar-ar} & \multicolumn{2}{c}{ar-en} & \multicolumn{2}{c}{es-es} & \multicolumn{2}{c}{es-en} & \multicolumn{2}{c}{tr-en}\\
\midrule
\contrastive & 72.4 & 72.2 & 72.7 & 74.2 & 79.7 & 81.0 & 71.7 & 72.0 & 77.2 & 77.0\\
\bitrans & 75.6 & 76.0 & 77.0 & 78.6 & 84.0 & 84.8 & 76.2 & 77.2 & 77.3 & 77.5\\
\mgtcontrastive & 73.2 & 73.1 & 73.5 & 75.1 & 80.2 & 81.3 & 72.4 & 72.4 & \bf 77.8 & \bf 78.0\\
\model & \bf 77.6 & \bf 78.1 & \bf 78.5 & \bf 78.8 & \bf 85.5 & \bf 85.7 & \bf 77.0 & \bf 77.4 & 77.0 & 77.0\\
\bottomrule
\end{tabular}
\end{center}
\caption{Full results on Cross-Lingual STS. We report results using both Pearson's $r \times 100$ and Spearman's $\rho \times 100$ across datasets, where Pearson's $r \times 100$ is the first column for each language pair and Spearman's $\rho \times 100$ is the second column.}
\label{tab:xlsts-full-results}
\end{table*}

For English semantic similarity, we use the SemEval semantic textual similarity (STS) tasks from 2012 to 2016~\cite{agirre-etal-2012-semeval,agirre-etal-2013-sem,agirre-etal-2014-semeval,agirre-etal-2015-semeval,agirre-etal-2016-semeval} as was done initially for sentence embeddings in~\cite{wieting-16-full}. As our test set, we report the average Pearson's $r$ over each year of the STS tasks from 2012-2016 as is convention in the top part of Table~\ref{tab:english-sts-full-results}. However, some recent work, like~\citet{reimers-gurevych-2019-sentence} computed Spearman's $\rho$ over concatenated datasets for each year of the STS competition. To be consistent with these works, we also include evaluations using this approach in the bottom part of Table~\ref{tab:english-sts-full-results}. One other difference between these two ways of calculating the results is the inclusion of the SMT dataset of the 2013 task. When computing the results using Pearson's $r$, this dataset is included, but when computing the results using Spearman's $\rho$, it is not included.

For cross-lingual semantic similarity and semantic similarity in non-English languages, we evaluate on the STS tasks from SemEval 2017. This evaluation contains Arabic-Arabic, Arabic-English, Spanish-Spanish, Spanish-English, and Turkish-English datasets. The datasets were created by translating one or both pairs of an English STS pair into Arabic (\texttt{ar}), Spanish (\texttt{es}), or Turkish (\texttt{tr}). Following convention, we report results with Pearson's $r$ for all systems, but also include results in Spearman's $\rho$ in Table~\ref{tab:xlsts-full-results}.

\subsection{Question Retrieval}

\begin{table}[htp]
\begin{center}
\small
\begin{tabular}{ lr } 
\toprule
Model & NQ \\
\midrule
\contrastive & 40.2\\
\bitrans & \bf 40.9\\ 
\mgtcontrastive & 40.3\\
\model & 40.8\\
\bottomrule
\end{tabular}
\end{center}
\caption{Full results on question retrieval on the NQ data. We evaluate retrieval accuracy $\times 100$ using PAQ as a question knowledge base.}
\label{tab:nq-full-results}
\end{table}

\begin{table*}[p!]
\begin{center}
\small
\begin{tabular}{ lrrrrrrrrrrrrr } 
\toprule
Model & \multicolumn{12}{c}{MKQA} \\
Language & ar & da & de & en & es & fi & fr & he & hu & it & ja & km \\
\midrule
\contrastive & 21.4 & 30.4 & 29.2 & 33.2 & 30.4 & 27.7 & 30.0 & 24.4 & 26.9 & 29.4 & 24.2 & 23.6\\
\bitrans & 19.4 & 30.8 & 30.5 & 29.8 & 28.7 & 29.7 & 28.0 & 30.3 & 27.5 & 27.9 & 26.2 & 23.4\\
\mgtcontrastive & \bf 24.6 & 32.0 & 30.5 & \bf 33.4 & \bf 31.6 & 29.8 & \bf 30.9 & 27.9 & 28.9 & \bf 31.0 & 27.3 & 24.9\\
\model & 21.4 & \bf 32.1 & \bf 32.5 & 31.5 & 30.3 & \bf 31.3 & 30.0 & \bf 31.9 & \bf 29.9 & 30.0 & \bf 29.9 & \bf 25.9\\
\midrule
\midrule
Language & ko & ms & nl & no & pl & pt & ru & sv & th & tr & vi & zh \\
\midrule
\contrastive & 22.0 & 30.5 & 29.4 & 33.2 & 30.4 & 27.7 & 30.0 & 24.8 & 27.5 & 29.6 & 24.6 & 24.2\\
\bitrans & 19.4 & 30.8 & 30.9 & 30.0 & 29.2 & 30.2 & 28.1 & 30.4 & 28.0 & 28.3 & 26.9 & 23.7\\
\mgtcontrastive & \bf 25.3 & 32.3 & 30.6 & \bf 33.4 & \bf 31.5 & 30.1 & \bf 31.0 & 28.1 & 29.9 & \bf 31.1 & 27.7 & 24.9\\
\model & 22.0 & \bf 32.4 & \bf 32.8 & 31.6 & 31.0 & \bf 31.6 & 30.4 & \bf 32.2 & \bf 30.4 & 30.2 & \bf 30.4 & \bf 26.2\\
\bottomrule
\end{tabular}
\end{center}
\caption{Full results on question retrieval on the MKQA data. We evaluate retrieval accuracy $\times 100$ using PAQ as a question knowledge base.}
\label{tab:mkqa-full-results}
\end{table*}

For our question retrieval evaluation, we report the accuracy (R@1) on the test sets of Natural Questions (NQ)~\cite{kwiatkowski-etal-2019-natural} shown in Table~\ref{tab:nq-full-results} and the Multilingual Knowledge Questions and Answers (MKQA)~\cite{longpre-etal-2021-mkqa} shown in Table~\ref{tab:mkqa-full-results}. We use the the Probably Asked Questions dataset (PAQ)~\cite{lewis-etal-2021-paq} as a knowledge base from which we look up the nearest neighbor of each question in the NQ and MKQA test sets using cosine similarity.

\subsection{Bitext Mining}

For bitext mining, we use the Tatoeba dataset introduced in \citet{artetxe-schwenk-2019-massively} and the 2018 Building and Using Parallel Corpora (BUCC) shared bitext mining task~\cite{zweigenbaum2018overview}. 

The Tatoeba dataset consists of 100-1000 pairs of data aligned to English for 112 languages. The accuracy for Tatoeba can be computed in two ways, depending if English is the target language or source language. We compute accuracy using cosine similarity in both directions for all 112 languages (19 are unseen in the training data) and average this score for all languages. 

The goal of the BUCC task is to find the gold {\it aligned} parallel sentences given two corpora (one being very large) in two distinct languages. Languages are aligned with English and consist of German (de), French (fr), Russian (ru), and Chinese (zh). Following \citet{schwenk-2018-filtering}, we evaluate on the publicly available BUCC data. This involves scoring all pairs between the source target sentences and finding the optimal threshold that separates the data. Using the threshold, we can compute the precision, recall, and $F_1$ of the alignments. We report $F_1 \times 100$ in our results.

We compare two different approaches for finding the sentence alignments. In the first, BUCC (cosine), we compute the cosine similarity between the non-English source sentences and the English target sentences, selecting the highest scoring English sentence as the match. In the second, BUCC (margin), we follow \citet{artetxe-schwenk-2019-margin} and use a margin-based scoring approach.

\begin{table*}[p!]
\begin{center}
\small
\begin{tabular}{ lrrrr|rrrrr } 
\toprule
Model & \multicolumn{4}{c}{Cosine} & \multicolumn{4}{c}{Margin} \\
\cmidrule{2-9} \vspace{-0.3cm} \\
& de & fr & ru & zh & de & fr & ru & zh \\
\midrule
\contrastive & 84.6 & 81.3 & 66.6 & 64.4 & \bf 96.2 & \bf 93.7 & \bf 92.1 & \bf 93.0\\
\bitrans & 90.1 & 85.5 & 84.1 & 84.1 & 93.6 & 90.3 & 91.3 & 92.4\\
\mgtcontrastive & 84.8 & 81.9 & 67.4 & 64.4 & 96.1 & 93.6 & \bf 92.1 & 92.9\\
\model & \bf 91.5 & \bf 86.8 & \bf 86.7 & \bf 86.1 & 94.3 & 91.0 & 91.8 & 92.8\\
\bottomrule
\end{tabular}
\end{center}
\caption{Full results on BUCC. We report results using both cosine similarity and the margin approach from ~\cite{artetxe-schwenk-2019-margin}. Results are reported as $F_1 \times 100$.}
\label{tab:bucc-full-results}
\end{table*}

\begin{table*}[p!]
\begin{center}
\small
\addtolength{\tabcolsep}{-4pt}
\begin{tabular}{llccccccccccccccccccccccc}
\toprule
Language & afr & amh & ang & ara & arq & arz & ast & awa & aze & bel & ben & ber & bos & bre & bul & cat\\
\midrule
\contrastive & \bf 97.6 & \bf 94.9 & 66.8 & \bf 95.0 & 61.8 & \bf 86.1 & \bf 91.3 & 74.0 & \bf 95.8 & 96.8 & 92.4 & 80.4 & \bf 97.5 & 47.2 & \bf 96.4 & \bf 97.8\\
\bitrans & 94.8 & 84.2 & 42.9 & 94.0 & 42.0 & 80.3 & 80.3 & 56.3 & 91.3 & 95.0 & 91.0 & 72.6 & 96.8 & 18.9 & 95.3 & 96.8\\
\mgtcontrastive & 97.4 & 93.5 & \bf 70.5 & 94.7 & \bf 64.2 & 85.7 & 90.9 & \bf 76.0 & \bf 95.8 & \bf 97.2 & \bf 92.8 & \bf 81.6 & 97.3 & \bf 47.9 & 96.2 & \bf 97.8\\
\model & 95.6 & 88.1 & 53.4 & 94.6 & 50.2 & 84.7 & 86.2 & 64.3 & 93.5 & 95.6 & 91.9 & 79.0 & 97.0 & 26.1 & 95.8 & 97.0\\
\midrule
\midrule
Language & cbk & ceb & ces & cha & cmn & cor & csb & cym & dan & deu & dsb & dtp & ell & epo & est & eus\\
\midrule
\contrastive & 86.1 & 62.3 & \bf 98.3 & \bf 44.9 & 97.5 & 35.1 & 67.8 & \bf 57.6 & \bf 97.3 & \bf 99.6 & 75.4 & 18.9 & \bf 97.4 & \bf 98.6 & 98.6 & 96.9\\
\bitrans & 78.0 & 48.8 & 97.3 & 33.6 & 95.6 & 18.4 & 48.6 & 37.0 & 96.0 & 99.3 & 53.1 & 8.3 & 95.8 & 98.3 & 97.8 & 94.8\\
\mgtcontrastive & \bf 86.9 & \bf 63.7 & 98.2 & 44.2 & \bf 97.7 & \bf 37.6 & \bf 69.8 & 56.4 & 97.2 & \bf 99.6 & \bf 76.9 & \bf 20.3 & 97.2 & 98.4 & \bf 98.8 & \bf 97.1\\
\model & 83.7 & 52.8 & 97.9 & 38.0 & 96.4 & 23.3 & 56.1 & 43.0 & 96.8 & 99.2 & 63.2 & 10.2 & 97.0 & 98.2 & 98.2 & 95.4\\
\midrule
\midrule
Language & fao & fin & fra & fry & gla & gle & glg & gsw & heb & hin & hrv & hsb & hun & hye & ido & ile\\
\midrule
\contrastive & 90.3 & 98.0 & 96.4 & \bf 88.2 & \bf 58.4 & \bf 80.4 & \bf 98.6 & 52.1 & \bf 93.8 & \bf 98.1 & \bf 98.5 & 80.1 & \bf 98.4 & \bf 96.2 & 93.0 & 92.8\\
\bitrans & 78.2 & 97.8 & 95.8 & 80.1 & 35.6 & 60.6 & 96.8 & 44.9 & 93.0 & 96.4 & 96.8 & 58.6 & 96.4 & 95.2 & 87.9 & 86.7\\
\mgtcontrastive & \bf 91.0 & \bf 98.2 & \bf 96.6 & 87.9 & 55.7 & 79.5 & \bf 98.6 & \bf 53.8 & 93.7 & 98.0 & 98.4 & \bf 82.4 & 98.2 & 96.1 & \bf 94.4 & \bf 93.1\\
\model & 82.6 & 98.0 & 96.0 & 83.5 & 39.4 & 62.7 & 97.4 & 50.4 & \bf 93.8 & 97.5 & 97.5 & 68.9 & 96.8 & 94.7 & 91.9 & 90.2\\
\midrule
\midrule
Language & ina & ind & isl & ita & jav & jpn & kab & kat & kaz & khm & kor & kur & kzj & lat & lfn & lit\\
\midrule
\contrastive & 96.8 & \bf 97.0 & 97.0 & 96.6 & 74.1 & 98.3 & 71.7 & 95.6 & \bf 92.8 & \bf 87.8 & 95.2 & \bf 76.3 & 17.4 & \bf 89.9 & 83.6 & \bf 98.2\\
\bitrans & 94.9 & 95.2 & 96.3 & 96.2 & 62.9 & 96.8 & 60.8 & 93.9 & 86.1 & 85.4 & 92.5 & 60.5 & 8.8 & 83.9 & 74.6 & 97.5\\
\mgtcontrastive & \bf 97.2 & 96.9 & \bf 97.1 & 96.6 & \bf 76.1 & \bf 98.6 & \bf 73.3 & \bf 96.2 & 92.3 & 87.5 & \bf 95.8 & 76.0 & \bf 17.8 & 89.8 & \bf 84.1 & 98.1\\
\model & 96.4 & 95.8 & 96.8 & \bf 96.8 & 69.5 & 97.2 & 67.3 & 95.8 & 87.7 & 86.3 & 93.5 & 67.7 & 11.7 & 86.5 & 79.0 & 97.8\\
\midrule
\midrule
Language & lvs & mal & mar & max & mhr & mkd & mon & nds & nld & nno & nob & nov & oci & orv & pam & pes\\
\midrule
\contrastive & 98.0 & \bf 98.5 & 94.5 & \bf 73.1 & \bf 30.8 & 97.6 & 94.4 & 90.5 & \bf 98.1 & 96.5 & \bf 98.5 & \bf 80.5 & 77.9 & 66.5 & 14.0 & 95.5\\
\bitrans & 97.0 & 98.2 & \bf 95.0 & 58.1 & 22.1 & 96.0 & 85.8 & 80.7 & 96.7 & 92.3 & 97.4 & 70.6 & 66.5 & 47.5 & 8.5 & 93.0\\
\mgtcontrastive & \bf 98.1 & \bf 98.5 & 94.6 & 72.5 & 29.5 & \bf 98.0 & \bf 95.0 & \bf 92.1 & 98.0 & \bf 97.0 & 98.3 & \bf 80.5 & \bf 78.0 & \bf 67.4 & \bf 14.6 & \bf 95.7\\
\model & 97.5 & 98.3 & \bf 95.0 & 63.6 & 26.8 & 96.4 & 89.8 & 84.5 & 97.3 & 93.5 & 97.6 & 76.3 & 72.1 & 55.9 & 9.9 & 94.5\\
\midrule
\midrule
Language & pms & pol & por & ron & rus & slk & slv & spa & sqi & srp & swe & swg & swh & tam & tat & tel\\
\midrule
\contrastive & 74.3 & \bf 99.0 & 95.9 & 98.0 & 95.3 & 98.0 & \bf 97.0 & 99.1 & \bf 98.6 & \bf 96.6 & \bf 97.5 & \bf 76.8 & 77.4 & 92.7 & \bf 92.0 & \bf 98.1\\
\bitrans & 62.1 & 97.2 & 95.7 & 97.6 & 95.0 & 97.4 & 96.3 & 98.8 & 98.1 & 95.6 & 96.8 & 48.7 & 67.1 & 90.9 & 82.8 & 96.2\\
\mgtcontrastive & \bf 76.9 & 98.9 & \bf 96.1 & \bf 98.1 & \bf 95.5 & \bf 98.2 & \bf 97.0 & \bf 99.2 & \bf 98.6 & 96.5 & \bf 97.5 & 73.2 & \bf 77.6 & \bf 92.8 & \bf 92.0 & 97.6\\
\model & 69.6 & 98.2 & 95.8 & 97.6 & 94.8 & 97.8 & 96.9 & 98.6 & 98.2 & 95.8 & 97.3 & 59.8 & 69.2 & \bf 92.8 & 86.2 & 97.4\\
\midrule
\midrule
Language & tgl & tha & tuk & tur & tzl & uig & ukr & urd & uzb & vie & war & wuu & xho & yid & yue & zsm\\
\midrule
\contrastive & 96.0 & 97.8 & 44.6 & \bf 98.9 & \bf 66.3 & 76.1 & 96.0 & \bf 95.4 & \bf 78.9 & 98.2 & 54.0 & 93.5 & 74.6 & 92.3 & \bf 94.1 & \bf 98.0\\
\bitrans & 91.7 & 97.0 & 30.3 & 98.2 & 43.8 & 54.4 & 95.2 & 91.8 & 64.3 & 97.4 & 33.3 & 88.7 & 59.5 & 82.8 & 90.8 & 96.0\\
\mgtcontrastive & \bf 96.4 & \bf 98.2 & \bf 47.5 & \bf 98.9 & 64.4 & \bf 77.6 & \bf 96.3 & \bf 95.4 & 78.3 & \bf 98.4 & \bf 54.5 & \bf 93.8 & \bf 75.7 & \bf 92.6 & \bf 94.1 & 97.8\\
\model & 93.0 & 97.5 & 38.7 & 98.8 & 56.7 & 64.1 & 95.5 & 93.5 & 68.7 & 97.9 & 37.7 & 91.0 & 63.7 & 86.0 & 93.0 & 96.6\\
\bottomrule
\end{tabular}
\end{center}
\caption{Full results on Tatoeba. We report results as accuracy $\times 100$.}
\label{tab:tatoeba-full-results}
\end{table*}

\section{Full Training Data} \label{sec:appendix-full-training-data}

We follow~\citet{artetxe-schwenk-2019-massively} in constructing our training data, sampling data from Europarl,\footnote{\url{http://opus.nlpl.eu/Europarl.php}}, United Nations~\cite{rafalovitch-dale-2009-united},\footnote{\url{https://opus.nlpl.eu/UN.php}} OpenSubtitles2018~\cite{lison-etal-2018-opensubtitles2018},\footnote{\url{http://opus.nlpl.eu/OpenSubtitles.php}}, Global Voices,\footnote{\url{https://opus.nlpl.eu/GlobalVoices.php}} Tanzil,\footnote{\url{https://opus.nlpl.eu/Tanzil.php}} and Tatoeba v2021-07-22.\footnote{\url{https://opus.nlpl.eu/Tatoeba.php}}

The only deviation from their data sampling approach is that we take care to not include any Tatoeba test data in our training data. Our final corpus has nearly 216 million training examples, slightly less than 220 million reported in~\citet{artetxe-schwenk-2019-massively}. We use both English and Spanish as pivot languages, so each pair includes at least one English or Spanish sentence, and attempt to use approximately the same amount of data for each language if possible. We note that we only have training data for 92 languages instead of the 93 in~\citet{artetxe-schwenk-2019-massively} due to not having training data for Aymara (ay). The full amount of English and Spanish parallel data used for each of the \N languages is reported in Table~\ref{tab:full-training-data}.

\begin{table*}[p!]
\begin{center}
\small
\addtolength{\tabcolsep}{-4pt}
\begin{tabular}{llccccccccccccccc}
\toprule
Language & af & am & ar & ay & az & be & ber & bg\\
\midrule
Training Pairs & 77,772 & 101,613 & 7,907,914 & 0 & 291,925 & 6,330 & 142,061 & 4,834,661\\
\midrule
\midrule
Language & bn & br & bs & ca & cbk & cs & da & de\\
\midrule
Training Pairs & 1,148,461 & 34,472 & 4,166,739 & 895,940 & 1,623 & 5,429,060 & 7,767,119 & 8,707,293\\
\midrule
\midrule
Language & dtp & dv & el & en & eo & es & et & eu\\
\midrule
Training Pairs & 1,064 & 98,320 & 6,601,989 & 4,913,379 & 447,622 & 4,913,379 & 5,093,003 & 1,432,979\\
\midrule
\midrule
Language & fi & fr & ga & gl & ha & he & hi & hr\\
\midrule
Training Pairs & 7,785,493 & 8,935,842 & 1,112 & 391,824 & 134,775 & 4,046,554 & 358,907 & 3,911,368\\
\midrule
\midrule
Language & hu & hy & ia & id & ie & io & is & it\\
\midrule
Training Pairs & 5,256,214 & 8,194 & 12,048 & 4,326,151 & 2,445 & 3,181 & 2,712,556 & 8,468,538\\
\midrule
\midrule
Language & ja & ka & kab & kk & km & ko & ku & kw\\
\midrule
Training Pairs & 3,981,886 & 360,136 & 26,460 & 6,172 & 3,266 & 2,566,495 & 98,733 & 3,463\\
\midrule
\midrule
Language & kzj & la & lfn & lt & lv & mg & mhr & mk\\
\midrule
Training Pairs & 614 & 27,515 & 6,096 & 3,629,769 & 2,119,995 & 537,953 & 69 & 4,037,896\\
\midrule
\midrule
Language & ml & mr & ms & my & nb & nds & nl & oc\\
\midrule
Training Pairs & 867,026 & 52,340 & 3,288,492 & 4,802 & 9,694 & 6,263 & 8,346,102 & 730\\
\midrule
\midrule
Language & pl & ps & pt & ro & ru & sd & si & sk\\
\midrule
Training Pairs & 5,407,190 & 32 & 8,276,190 & 4,814,046 & 9,416,934 & 98,412 & 1,016,660 & 5,094,752\\
\midrule
\midrule
Language & sl & so & sq & sr & sv & sw & ta & te\\
\midrule
Training Pairs & 5,099,577 & 98,976 & 3,619,914 & 3,977,191 & 7,680,683 & 201,379 & 150,023 & 42,877\\
\midrule
\midrule
Language & tg & th & tl & tr & tt & ug & uk & ur\\
\midrule
Training Pairs & 135,245 & 3,849,777 & 34,829 & 5,854,059 & 132,273 & 101,989 & 1,687,685 & 844,052\\
\midrule
\midrule
Language & uz & vi & wuu & yue & zh\\
\midrule
Training Pairs & 148,860 & 3,905,401 & 929 & 4,525 & 7,636,488\\
\bottomrule
\end{tabular}
\end{center}
\caption{Full training data for each language. The total number of pairs is the sum of using English and Spanish as pivot languages.}
\label{tab:full-training-data}
\end{table*}